\documentclass[letterpaper, 10 pt, conference]{ieeeconf}  

        \IEEEoverridecommandlockouts                              
        
        \usepackage{epsfig} 
        \usepackage{algorithm}
        \usepackage{algorithmic}
        \usepackage{amsmath,graphicx}                                                      
        \usepackage{amsfonts}
        \usepackage{color}
        \usepackage{xfrac}
        \usepackage{caption}
        \usepackage{subfigure}
        \usepackage{amssymb}
        \usepackage{bm}

        \title{\LARGE \bf
        Where Should We Place LiDARs on the Autonomous Vehicle? \\- An Optimal Design Approach
        }

        \author{Zuxin Liu$^{1}$, Mansur Arief$^{2}$ and Ding Zhao$^{2,*}$
        \thanks{* Corresponding author: Ding Zhao. Email: ({\tt\small dingzhao@cmu.edu})}%
        \thanks{$^{1}$Zuxin Liu is with the School of Instrumentation Science and Opto-electronics Engineering, Beihang University, China}%
        \thanks{$^{2}$Mansur Arief and Ding Zhao are with the Department of Mechanical Engineering, Carnegie Mellon University, USA.}%
        }

\begin{document}
        \maketitle
        \thispagestyle{empty}
        \pagestyle{empty}
        
        \begin{abstract}
                Autonomous vehicle manufacturers recognize that LiDAR provides accurate 3D views and precise distance measures under highly uncertain driving conditions. Its practical implementation, however, remains costly. This paper investigates the optimal LiDAR configuration problem to achieve utility maximization. We use the perception area and non-detectable subspace to construct the design procedure as solving a min-max optimization problem and propose a bio-inspired measure -- volume to surface area ratio (VSR) -- as an easy-to-evaluate cost function representing the notion of the size of the non-detectable subspaces of a given configuration. We then adopt a cuboid-based approach to show that the proposed VSR-based measure is a well-suited proxy for object detection rate. It is found that the Artificial Bee Colony evolutionary algorithm yields a tractable cost function computation. Our experiments highlight the effectiveness of our proposed VSR measure in terms of cost-effectiveness configuration as well as providing insightful analyses that can improve the design of AV systems.
        \end{abstract}

        \section{INTRODUCTION}
        \label{sec:intro}
        Critical to establishing safe driving for an autonomous vehicle (AV) is ensuring how rapidly and accurately it can perceive the surrounding environment and plan for maneuvers. Extensive efforts undertaken to improve AVs' perception include installing industry-grade cameras, such as Grashopper2, Flea2, etc. \cite{maddern2014illumination, geiger2013vision, hane20173d}. These efforts are highly dependent on the device configurations themselves, but even more on external environmental conditions, such as lighting. Therefore, many AV manufacturers have embraced LiDAR because it provides distance measures, thus reconstructing 3D views, and gives 360-degree coverage up to a reasonable distance. Some AV designs even incorporate multiple, expensive LiDARs to gain an even higher degree of ‘seeing’ capability based on sensor redundancy \cite{schwarz2010lidar}. Unfortunately, the market price of LiDARs remains substantial \cite{jayaweera2018autonomous}. For example, a 16-beam Velodyne LiDAR costs almost \$8,000 \cite{velod16}.
        
        
        
        
        To the best of our knowledge, the research on optimal, lower-cost LiDAR configurations for AVs is rare. A few studies have investigated how to achieve optimal configuration for the camera or 3D sensors, e.g., \cite{dybedal2017optimal} proposed an optimal camera configuration method to achieve the largest field of view, while \cite{banta1996autonomous} optimize the camera's pose for motion capture systems. Both methods cannot be directly applied for optimal LiDAR configuration, because, compared to cameras, the receptive field of LiDAR is much more discrete. Previously, in \cite{mou2018optimal}, we studied the LiDAR configuration problem by considering it as a min-max optimization and using a cylindrical representation as a proxy to the cost function and a mixed integer programming (MIP) to solve the model. 
        
        Motivated by the fact that our previous work suffers from the curse of dimensionality, i.e., it is unsuitable for solving large-scale problems, such as those involving multiple LiDARs, in this paper, we propose to reformulate the cost function by utilizing an easy-to-solve proxy. We use volume to surface area ratio (VSR), which has both geometric and bionic interpretations as a measure to evaluate the performance of a particular configuration. The resulting solution allows us to analyze the trade-offs between an AV's ’seeing’ capability and a LiDAR's design cost. Our goal is to find the right balance between the reliability and affordability of AV perception systems without sacrificing driving safety. Our proposed approach also allows us to tractably evaluate the complex sensors specifically designed for autonomous systems, such as artificial compound eyes \cite{floreano2013miniature, floreano2015science}.
        
        \begin{figure*}[ht]
                \centering
                \subfigure[Cone-like Perception area of LiDAR beams]
                { 
                        \label{fig:cone} 
                        \includegraphics[width=2.15in]{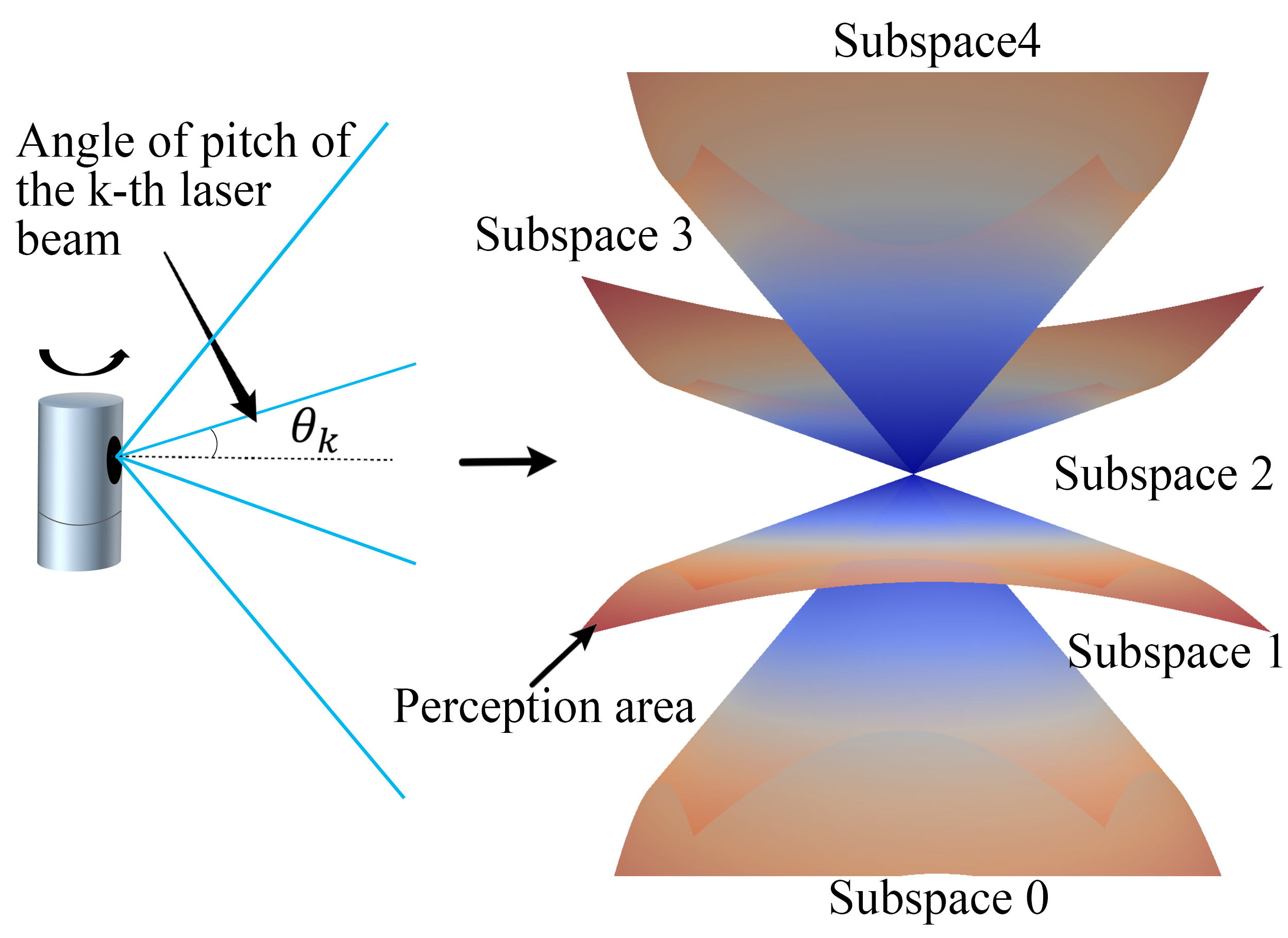}
                }  
                \subfigure[Region of interest, where the red, purple, green represent 3 different LiDARs' beams]
                { 
                        \label{fig:ROI} 
                        \includegraphics[width=2.15in]{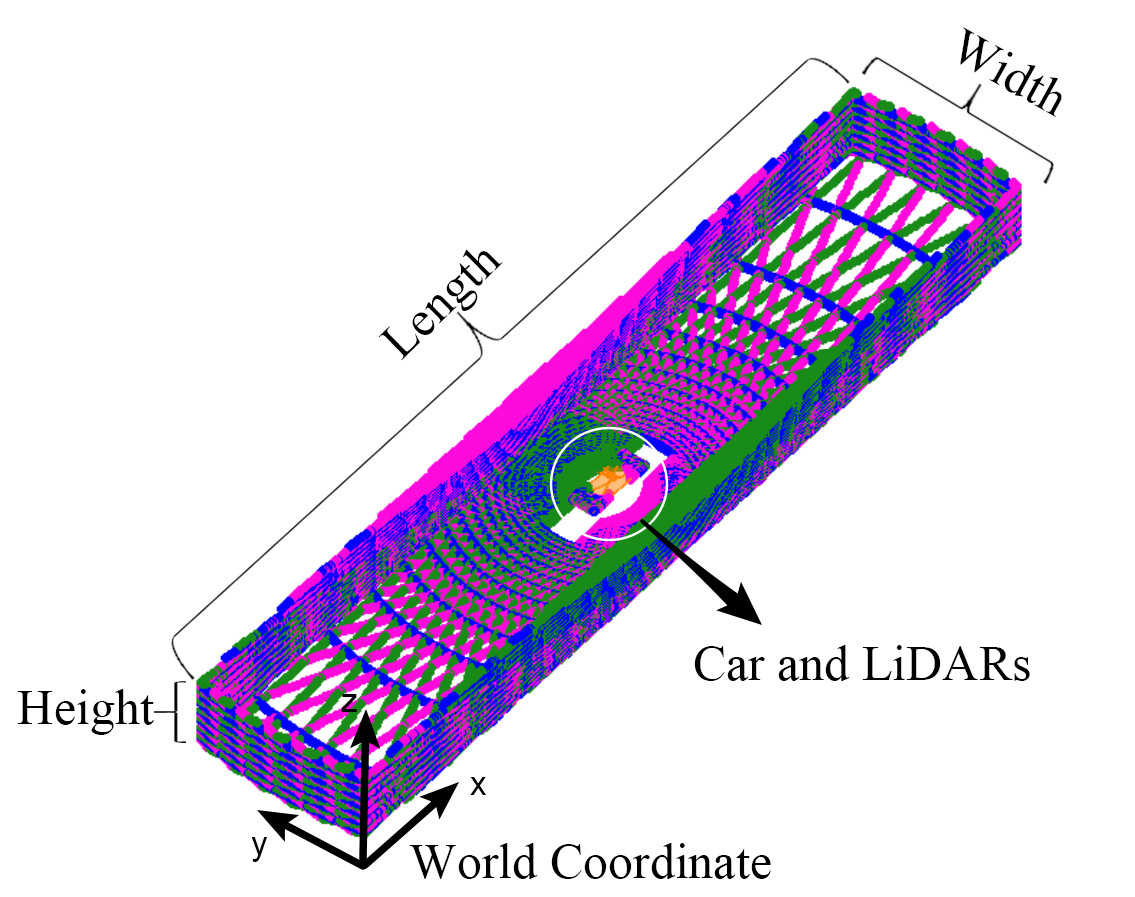}
                } 
                \subfigure[Non-detectable subspace] 
                {  
                        \label{fig:sphere} 
                        \includegraphics[width=2.15in]{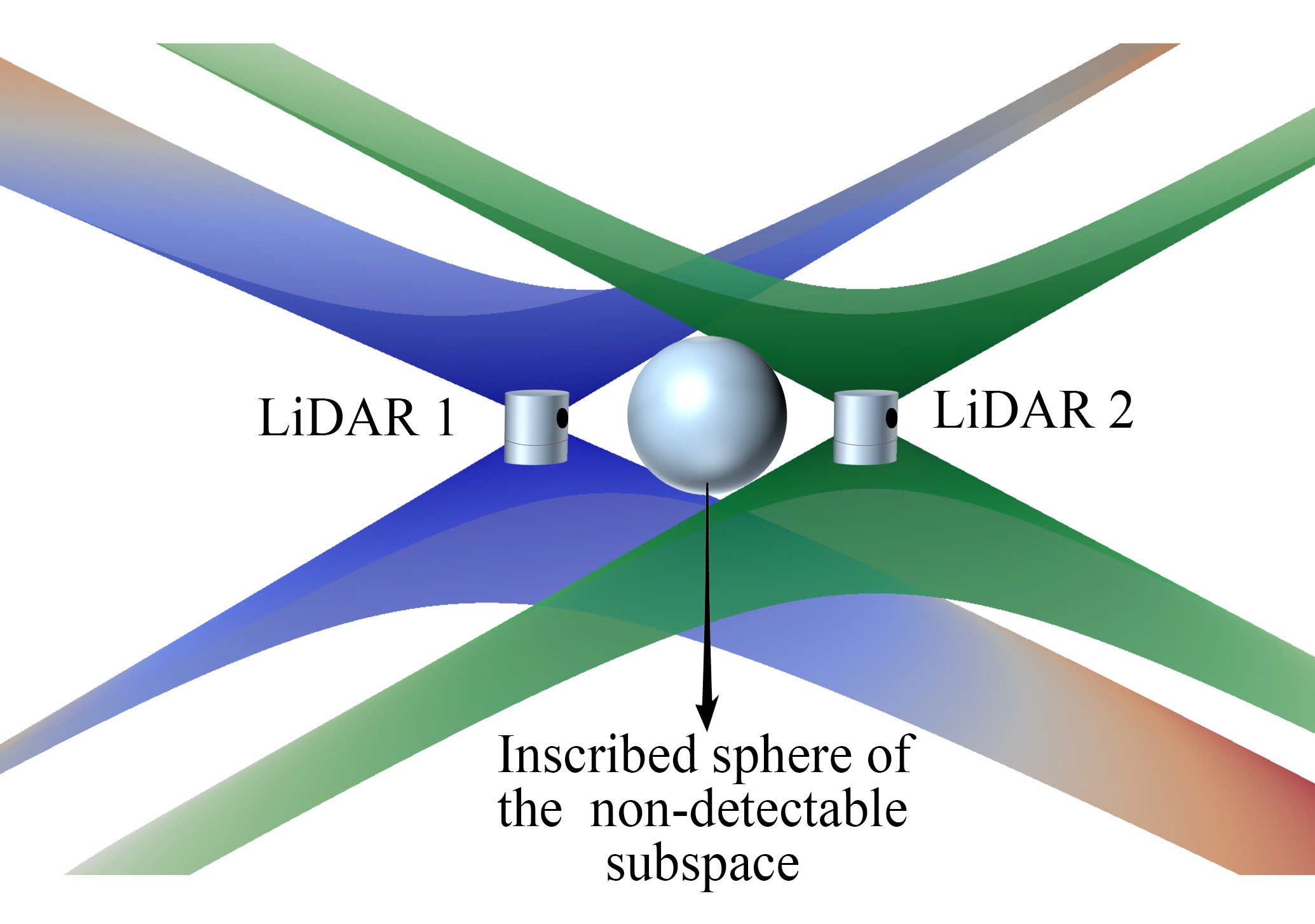}
                }
                \caption{ Perception area, ROI and subspace demonstration}
                \label{fig:subspace}
        \end{figure*}

        The remainder of this paper is as follows. Section \ref{sec:method} introduces the general concept and structure of our proposed approach to the optimal LiDAR configuration problem. Section \ref{sec:formulation} describes the model and our use of the Artificial Bee Colony (ABC) algorithm. Section \ref{sec:experiment} explains the experiment settings and discusses the results. Section \ref{sec:conclusion} concludes and suggests some future directions.
        
        \section{Methodological Framework}
        \label{sec:method}

        In this section, we begin by defining the region of interest (ROI) and the perception area of LiDAR beams, both of which form the basis of the configuration problem. Then we introduce the non-detectable subspace and propose the maximum VSR as the cost function. We also adopt the Artificial Bee Colony (ABC) algorithm to solve the optimization problem.
        
        \subsection{ROI and perception area of LiDAR beams}
        We account for the limitation of LiDAR's detection range by considering that the LiDAR's placement does not influence the regions that are far away from the AV. This consideration aligns with many real-world applications where the perception algorithm chooses only a portion of the LiDAR sensor data as its input. For example, VoxelNet \cite{zhou2017voxelnet} use the point cloud data within a certain bounding box around the LiDAR as the input to perform the object detection task. Considering the ROI around the AV in the LiDAR configuration problem, we want to obtain as much information as possible by well-covering the whole ROI with LiDAR beams.

        We define the ROI by three-dimensional geometric properties -- the length, width, and height -- of a cuboid in the world coordinate. Fig. \ref{fig:ROI} shows an example of an AV located at the bottom-center of the $x$-$y$ plane of the ROI, which allows us to define the perception area as the whole trajectory of LiDAR beams after a 360-degree rotation, i.e., each beam forms a surface. Note that we can only detect the objects that intersect with the perception area.
        
        To simplify the representation, we assume that the beam always provides a perception line along its trajectory, i.e., the trajectory line forms a cone when the LiDAR beams rotate 360-degrees. Thus, we can model the whole perception area of a multi-beams LiDAR as the union of finitely many cones sharing the same vertex and vertical axis. Fig. \ref{fig:cone} shows the cone-like perception area of the LiDAR beams, where $\theta_k$ represents the pitch angle of the $k$-th laser beam. 
        
        

        \subsection{Non-detectable subspace segmentation}
        Given the ROI and the LiDAR configuration, the perception area of each LiDAR can intersect with each other as well as with the ROI boundaries. Therefore, the perception area segments the ROI into many irregular shape subspaces. We define them as non-detectable subspaces because the object located within one of the subspaces cannot be detected if both the LiDAR and object are static.
        
        Fig. \ref{fig:sphere} shows an example of the non-detectable subspaces. We let $N_l$ denote the number of LiDARs to be placed. Each LiDAR has $N_b$ beams. Fig. \ref{fig:cone} shows one LiDAR segmenting the ROI into $N_b+1$ subspaces. From bottom to top, we denote each subspace by a number in ascending order from 0 to $N_b$. Therefore, for $N_l$ LiDARs, the maximum number of subspace $N_s$ is
        \begin{equation}
                N_s=(N_b+1)^{N_l}.
        \end{equation}
        We note that it is hard to represent the non-detectable subspaces analytically because their shapes are different and irregular. Section \ref{sec:formulation} introduces our solution, which is a cuboid-based segmentation method.
        
        \subsection{Cost function}
        \label{subsec:cost}
        Note that all of the non-detectable subspaces should be \textit{as small as possible} to obtain the most informative perception from the LiDAR. Since the notions of how big or small are somewhat vague because each shape is different and irregular, we use the radius of the inscribed sphere to define the size of the subspace. As Fig. \ref{fig:sphere} shows that the subspace of the largest size, i.e., the one with the largest ‘blindspot’, is the one with the largest inscribed sphere. 
        
        To fix the worst possible case, we express the optimal LiDAR configuration as a min-max optimization problem to minimize the largest inscribed spheres from all subspaces. See \cite{mou2018optimal} for the details. We note, however, that it is very hard to solve the inscribed sphere directly, so we propose another easy-to-solve indicator -- the volume to surface area ratio (VSR) -- which is inspired by bionics, to define the size of the subspace.
                   
        \begin{figure}
                \centering
                \subfigure[Segment ROI into cuboids]
                { 
                        \label{fig:roi_cubes} 
                        \includegraphics[width=1.55in]{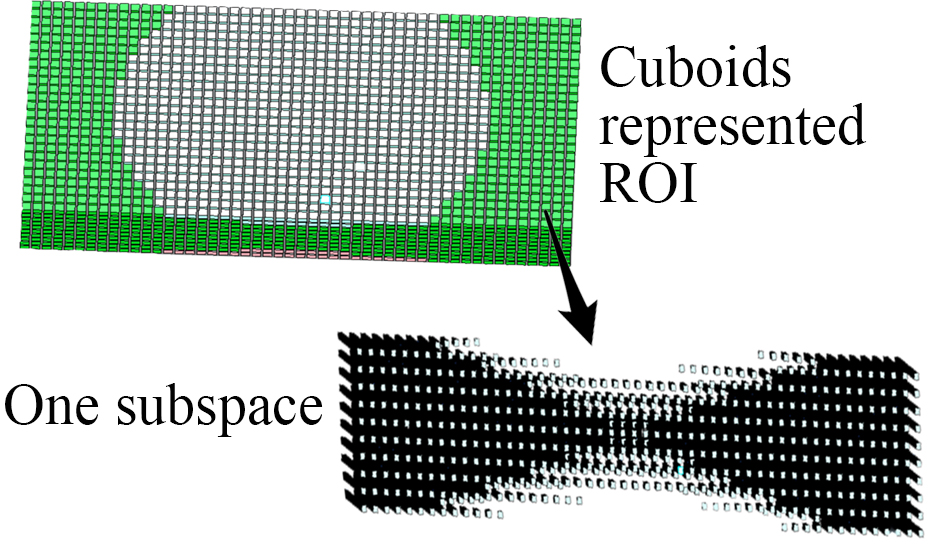}
                } 
                \subfigure[BFS-based second-level segmentation result]
                {
                        \label{fig:bfs} 
                        \includegraphics[width=1.55in]{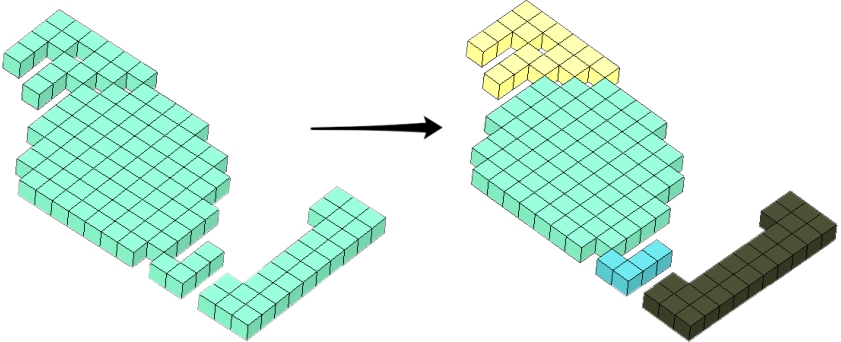}
                }
                \caption{Subspace segmentation}
                \label{fig:nothing}
        \end{figure}
        
        In the bionic world, VSR could be one dominating factor in the shapes and sizes of cells \cite{lewis1976surface,harris2018surface}. A cell receives nutrition from the outside environment through the cell membrane; the surface area represents the cell's efficiency in absorbing nutrition, and its volume represents the amount of nutrition needed. As the cell grows bigger, its volume and membrane surface area also enlarge. Because the increase rate of the surface area is less than the volume, and the rate of absorption decreases, it is not economical for a cell to grow too big \cite{lewis1976surface}.
        
        In this paper, we consider each non-detectable subspace as a cell and each surface area as the membrane. Loosely speaking, we consider the object to be detected as nutrition. Thus, we want to obtain the smallest possible VSR for optimal detection, especially of small objects. Specifically, we are interested in mitigating the worst-case scenario, i.e., we want to minimize the maximum VSR from the constructed non-detectable subspaces of the ROI. 
        
        We can also describe the VSR by using the formula of the radius of an inscribed sphere in a polyhedron \cite{insphere}. When we know the volume, denoted as $V$, and the surface area of the polyhedron, denoted as $S$, we can express the radius of inscribed sphere $R$ as $R=3 \times \frac{V}{S}$. Therefore, the VSR can represent the `size' of any shape of our interest.
        
\begin{figure}
                        \centering
                 \includegraphics[width=\linewidth]{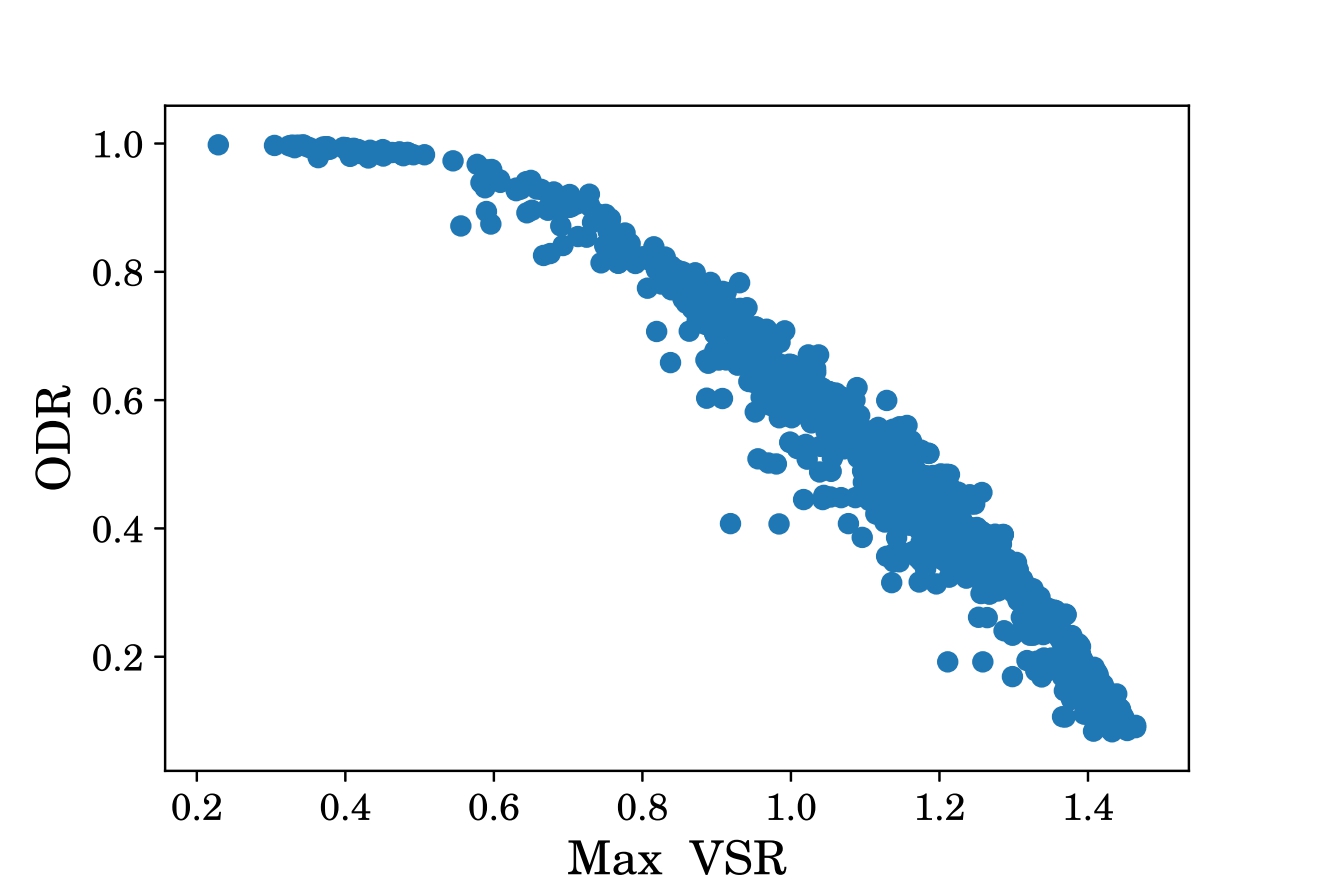}
                        \caption{General relationship between VSR and ODR}
                        \label{fig:vsr-odr}
                \end{figure}

        In practice, engineers value the ability of LiDAR configurations to detect objects. In this paper, the object detected rate (ODR) represents the probability of an AV to detect a certain object. We let some objects of interest, which can also be represented as the cuboids-based objects, randomly appear within the ROI. We denote $N_{ds}$ as the number of subspaces occupied by the cuboids. If $N_{ds}> thres$, where $thres$ is a predefined threshold, we assume the object is detected. In the following experiment, we simulate the object to randomly appear within the ROI $M$ times and tally the occurrences $T$ where the object is detected. This setting allows us to calculate the ODR as $ODR=\frac{T}{M}$.
        Minimizing the maximum VSR will increase ODR (Fig. \ref{fig:vsr-odr} is the general relationship between these measures). Therefore, we set the objective of the optimization model as minimization.
        
        \subsection{Artificial Bee Colony optimization algorithm}
        \label{sec:abc}
        We note that the optimization problem is non-convex, and the solution space is continous, high-dimensional and constrained. Therefore, using heuristic optimization method is a suitable choice in this case. We adopt the Artificial Bee Colony (ABC) optimization algorithm \cite{karaboga2007powerful} as the solver for this LiDAR configuration problem, because its better performance and fewer hyper-parameters than other similar population-based algorithms \cite{karaboga2009comparative}.
        
        The ABC optimization algorithm is inspired by the behavior of foraging bee swarms. The solutions correspond to food sources, and the cost function corresponds to the nectar amount (i.e., the fitness) of a particular food source. Greedy selection policy is employed to ensure that the algorithm converges quickly, and a roulette wheel selection method is adopted to explore around the food sources for the nectar of the highest amount.
        
        Technically speaking, the swarm of bees is divided into three groups: employed bees, onlookers, and scouts. The employed bees measure the nectar amount of food sources, and each employed bee corresponds to a food source. The onlookers adopt the roulette wheel selection method to select the existing food sources to be further explored. If the quality of an existing food source has not been improved within a certain number of iterations, its corresponding employed bee abandons the food source and becomes a scout to search for a new food source randomly. See \cite{karaboga2014comprehensive} for a comprehensive review of ABC algorithms. 
        
        
        
        \section{Model Formulation}
        \label{sec:formulation}
        As mentioned, it is very hard to represent the subspaces analytically, so we adopt a cuboid-based method for approximation. By segmenting (discretizing) the ROI into many small cuboids, we only need to determine which subspaces belong to which cuboids. Fig. \ref{fig:nothing} shows the steps: segment the ROI into many cuboids; transform each cuboid from world global coordinate to LiDAR local coordinate; perform first-level segmentation and use base-n notation to represent each subspace, and for each base-n notation use breadth-first search method to perform second-level segmentation. 
        
        \subsection{Cuboid coordinate transformation}
        We denote $\theta_k$ as the $k$-th laser beam of a LiDAR, and formulate the cone-like perception area as:
        \begin{equation}
                z_p=\tan \theta_{k} \sqrt{x_p^2+y_p^2},
        \end{equation}
        where $x_p,y_p,z_p$ are the coordinate points on the perception area relative to the LiDAR coordinate system.
        Since it is difficult to formulate the cone in the world coordinate if the LiDAR rotates, we transform the cuboid from world coordinate into LiDAR coordinate to facilitate the following segmentation.
        
        Let $[x_w,y_w,z,\alpha _w,\beta _w,\gamma _w]$ denote the pose of a LiDAR in the world coordinate, where $[x_w,y_w,z]$ represent the position and $[\alpha _w,\beta _w,\gamma _w]$ represent the orientation. Then we obtain the rotation matrix $\bm{R_{\alpha \beta \gamma}}$ as
        \begin{equation}
                \label{eq:transform}
                \bm{R_{\alpha \beta \gamma}} = \left[
                        \begin{matrix}
                                c_\alpha c_\beta & c_\alpha s_\beta s_\gamma-s_\alpha c_\gamma & c_\alpha s_\beta c_\gamma -s_\alpha s_\gamma \\
                                s_\alpha c_\beta & s_\alpha s_\beta s_\gamma-c_\alpha c_\gamma & s_\alpha s_\beta c_\gamma -c_\alpha s_\gamma\\
                                -s_\beta & c_\beta s_\gamma & c_\beta c_\gamma
                                \end{matrix}
                                \right] ,
        \end{equation}
        where $c$ represents $\cos$ and $s$ represents $\sin$ in short. For example, $c_\alpha$ represents $\cos(\alpha)$. We denote $\bm{T_{xyz}}=[x,y,z]^T$ as the translation vector, and express the transformation matrix from the LiDAR coordinate to the world coordinate $\bm{T_{wl}}$ as
        \begin{equation}
                \bm{T_{wl}} = \left[
                        \begin{matrix}
                                \bm{R_{\alpha \beta \gamma}} & \bm{T_{xyz}} \\
                                \bm{0}^T & 1
                                \end{matrix}
                                \right]. 
        \end{equation}
        Finally, we denote $\bm{X_{w}} = [x_{w},y_{w},z_{w}]$ and $\bm{X_{l}} = [x_{l},y_{l},z_{l}]$ as the coordinates of the cuboids in the world coordinate system and LiDAR coordinate system, respectively. Then, the transformation between the coordinate system satisfies
        \begin{equation}
                \bm{X_{l}}= \bm{T_{wl}}^{-1} \bm{X_{w}}.
        \end{equation}
        The two systems, which represent the cuboids in both the LiDAR and the world systems, set the stage for the two segmentation processes, which we decompose into two levels: base-n first-level segmentation and breadth-first search (BFS) second-level segmentation.
        
        \subsection{Base-n notation first-level segmentation}
        \label{subsec:base-n}
        After obtaining the cuboid coordinates in the LiDAR coordinate, the next step is to determine which subspace the cuboid belongs to. Here, we use base-n notation to identify the subspace.

        We assume there are $N_l$ $N_b$-beam LiDARs to be placed. According to equation (1), each LiDAR could segment the ROI into $N_b+1$ subspaces, resulting in $(N_b+1)^{N_l}$ maximum number of subspaces. Therefore, each subspace can be represented by a base-n notation $d_1d_2...d_i...d_n$, where $n$ equals the LiDARs' number $N_l$, and $d_i$ indicates the number of subspaces of the $i$-th LiDAR, ranging from 0 to $N_b$.
        
        We denote $\theta_{ik}$ as the pitch angle of the $k$-th beam of the $i$-th LiDAR, where $k \in {0,1,...,N_b}$. $ [x_{il},y_{il},z_{il}]$ is the cuboid coordinates in the $i$-th LiDAR coordinate system. To determine the digit $d_i$, we need to find the unique beam $k$ that satisfies 
        \begin{equation}
        \left\{
                \label{eq:rule}
                        \begin{array}{lr}
                        z_{il} \geq \tan \theta_{i(k-1)} \sqrt{x_{il}^2+y_{il}^2},  &  \\
                        &  1 \leq k \leq N_b-1  \\
                        z_{il} < \tan \theta_{ik} \sqrt{x_{il}^2+y_{il}^2} &  
                        \end{array}
        \right.
        \end{equation}
        or
        \begin{equation}        
                z_{il} < \tan \theta_{ik} \sqrt{x_{il}^2+y_{il}^2}, \quad \quad  k= 0 \\
        \end{equation}
        or
        \begin{equation}        
        z_{il} \geq \tan \theta_{ik} \sqrt{x_{il}^2+y_{il}^2},  \quad \quad k= N_b \\
        \end{equation}

        Then, the digit $d_i=k$ and thus, iterating over all the LiDARs allow us to determine the cuboids' subspaces in the ROI.
        
        \subsection{Breadth-first search (BFS) based second-level segmentation}
        A base-n notation of the subspace denotes each cuboid in the ROI. The cuboids of the same base-n notation are the non-detectable subspace segmented by the laser beams. However, the segmentation is not thorough. For example, the right side of Fig. \ref{fig:bfs} shows the cuboids of the same base-n notation. Therefore, we use a BFS-based method to perform the second-level segmentation.
        
        We classify the cuboid that has an overlapping face with the root node into the same group of the root cuboid. After searching all the non-visited candidate cuboids, we mark the root node cuboid as visited. We implement the same process for the left non-visited candidate cuboids until all the cuboids are marked as visited. The left side of Fig. \ref{fig:bfs} shows the result of performing the second-level segmentation.
        
        \subsection{Optimization model}
        To construct the optimization problem, we denote the number of LiDARs as $N_l$ and the number of laser beams as $N_b$. We denote the $i$-th LiDAR configuration as $C_i=[x_i,y_i,z_i,\alpha _i,\beta _i,\gamma _i]$, where $\alpha _i,\beta _i,\gamma _i$ each represents the yaw, pitch and roll of the LiDAR in world coordinate, respectively. 
        
        Since the LiDAR configuration $C$ is limited by the installation position of the car, we define $C_{\min}=[x_{\min},y_{\min},z_{\min},\alpha _{\min},\beta _{\min},\gamma _{\min}]$ and $C_{\max}=[x_{\max},y_{\max},z_{\max},\alpha _{\max},\beta _{\max},\gamma_{\max}]$ which represent some lower- and upper-bound of the LiDAR configurations, respectively. As such, the $i$-th configuration $C_i$ is valid if it satisfies the physical constraint:
        
        \begin{equation}
                C_{\min} \leq C_i \leq C_{\max}.  
        \end{equation}
        Let $CN=\{C_1,C_2,...,C_{N_l}\}$ be the configuration set. 
        Further, if we denote the VSR of the $j$-th subspace as $Loss_j (CN)$ for any given configuration set $CN$ and denote the set of cuboids coordinate in $j$-th subspace as $S_j$, then
        \begin{equation}
                Loss_j(CN)=\frac{V(S_j)}{SA(S_j)},   
        \end{equation}
        where $V(S_j)$ and $SA(S_j)$ are the volume and surface area of the $S_j$ cuboid set, respectively. Solving the $V(S_j)$, only requires calculating the number of the cuboids. We denote $[e_x, e_y, e_z]$ as the cuboid’s edges’ length along the $x,y,z$ axis such that the number of cuboids in $S_j$ is $n_j$. We solve $V(S_j)$ by:
        
        \begin{equation}
                V(S_j) = e_x e_y e_z n_j 
        \end{equation}
        
        The surface area $SA(S_j)$ comprises 3 parts: the surface area along $x-y$ plane $SA_{xy}$, along $x-z$ plane $SA_{xz}$ and along $y-z$ plane $SA_{yz}$. To know the size of the cuboid, we only need to count the overlap faces along the different planes.
        
        The algorithms to solve the 3 parts are the same, so here we use $SA_{xy}$ as an example. We denote $S$ as the cuboid set, and $S_{ix},S_{iy},S_{iz}$ as the $i-th$ cuboid's index along the $x,y,z$ axes. The steps are shown in Algorithm \ref{alg:SA}.
        
        \begin{algorithm}[htb] 
                \caption{ Solving the Surface Area of the Cuboid Set} 
                \label{alg:SA} 
                \begin{algorithmic}[1] 
                \REQUIRE ~ Cuboid set, $S$; Cuboid resolution, $[e_x, e_y, e_z]$;
                \ENSURE ~ The surface area along $x-y$ plane, $SA_{xy}$;
                \STATE Sort the coordinates of cuboids in $S$ according to $x-y-z$ axes in an increasing order
                \STATE $sz=Size(S)$ , $cnt=0$, $i=1$
        
                \REPEAT
                        \IF{$S_{ix} = S_{(i-1)x}$ and $S_{iy} = S_{(i-1)y}$ and $S_{iz} = S_{(i-1)z}+1$  }
                        \STATE $cnt=cnt+1$
                        \ENDIF
                \STATE $i=i+1$
                \UNTIL $i = sz$
                \STATE $SA_{xy}=2 (sz-cnt)e_x e_y$
                \end{algorithmic}
        \end{algorithm}

        We formulate the optimization problem as
        \begin{equation}
        \label{eq:optproblem}
                \begin{aligned}
                \overline{CN} = \arg & \min_{CN} \max_{j} & Loss_j(CN) \\
                &\textrm{subject to} & C_{\min} \leq CN \leq C_{\max}\ ,
                \end{aligned}
        \end{equation}
        where $\overline{CN}$ represents the optimized configuration solution.
        
        To solve (\ref{eq:optproblem}), we deploy the ABC algorithm and generate a swarm of artificial bees to explore the solution space.
        Let $x_i=(x_{i1},...,x_{i2},...,x_{id})$ ($i=1,2,…,\tau$) denote the solutions, where $\tau$ represents the number of bees and $d$ represents the solution’s dimension. $fit(x_i)$ represents the quality of the solution $x_i$, which is negatively related with our cost function. Defining probability $P_i$ as
        \begin{equation}
        \label{eq:onlook}
                P_i=\frac{fit(x_i)}{\sum_{i=1}^{\tau}fit(x_i)},
        \end{equation}
        allows the onlookers to select the solution $x_i$ and examine its quality $fit(x_i)$. Next, the algorithm prescribes a new solution $v_i$ near the currently selected solution $x_i$ by computing
        \begin{equation}
        \label{eq:moveon}
                v_{ij}=x_{ij}+\phi_{ij}(x_{ij}-x_{kj}),
        \end{equation}
        where $ k \in \{ 1,2,...,\tau\}$ and $ j \in \{ 1,2,...,d\}$ are randomly chosen indices. We note that $\phi_{ij}$ is a random number in the range $[-1,1]$. If the $fit(v_{i})>fit(x_{i})$, then we replace the solution $x_i$ with $v_i$, i.e., the bees move to $v_i$. Adopting the analogy described in Section \ref{sec:abc}, Algorithm \ref{alg:Framwork} summarizes the iterative steps of the ABC algorithm.
        
        \begin{algorithm}[htb] 
                \caption{ Artificial Bee Colony (ABC) Algorithm} 
                \label{alg:Framwork} 
                \begin{algorithmic}[1] 
                \REQUIRE ~Number of bees, $\tau$; Iteration number, $IterMax$; Threshold to abandon a food source, $thres$;
                \ENSURE ~ The optimal solution, $x_i$, $i=1,...,\tau$;
                \STATE Randomly initialize the food source $x_i$, $i=1,...,\tau$
                \STATE Get the quality of each food source $fit(x_i)$
                \STATE Set $cycle=0$ and $cnt_i=0$, $i=1,...,\tau$
                \REPEAT
                \FOR{each food source $x_i$}
                        \STATE Generate new food source $v_i$ using equation (\ref{eq:moveon})
                        \IF{$fit(v_i)>fit(x_i)$}
                        \STATE Replace $x_i$ with $v_i$
                        \ELSE 
                        \STATE $cnt_i=cnt_i+1$
                        \ENDIF
                \ENDFOR
                \FOR{each onlooker}
                        \STATE Select a food source using a roulette wheel selection method based on equation (\ref{eq:onlook})
                        \STATE Generate new food source $v_i$ around $x_i$ using equation (\ref{eq:moveon})
                        \IF{$fit(v_i)>fit(x_i)$}
                        \STATE Replace $x_i$ with $v_i$
                        \ELSE 
                        \STATE $cnt_i=cnt_i+1$
                        \ENDIF
                \ENDFOR
        
                \FOR{each food source $x_i$}
                \IF{$cnt_i=thres$}
                \STATE Replace $x_i$ with a randomly generated solution
                \ENDIF
                \ENDFOR
                \STATE $cycle=cycle+1$
                \UNTIL $cycle$ equals $IterMax$
                \end{algorithmic}
        \end{algorithm}
        
        \section{Experiment and Discussion}
        \label{sec:experiment}

        We consider an ROI of size $[60, 20, 4]$ in meters, excluding the rectangular region $x \in [27,33], y \in [8,12], z \in [0,4]$, which is very close to the AV and we do not consider its in the perception algorithm at this point. The lower- and upper- bound of the LiDARs’ pose are $[28, 9, 2.2, 0, 0, 0]$ and $[31, 11, 3, 3.1415, 3.1415, 0]$, where the first three components are measured in meters and the last three are measured in radians. The $yaw$ angle is not be optimized because the LiDAR will rotate 360 degrees. 
        

        \begin{figure}
                \centering
                \includegraphics[width=2.51in]{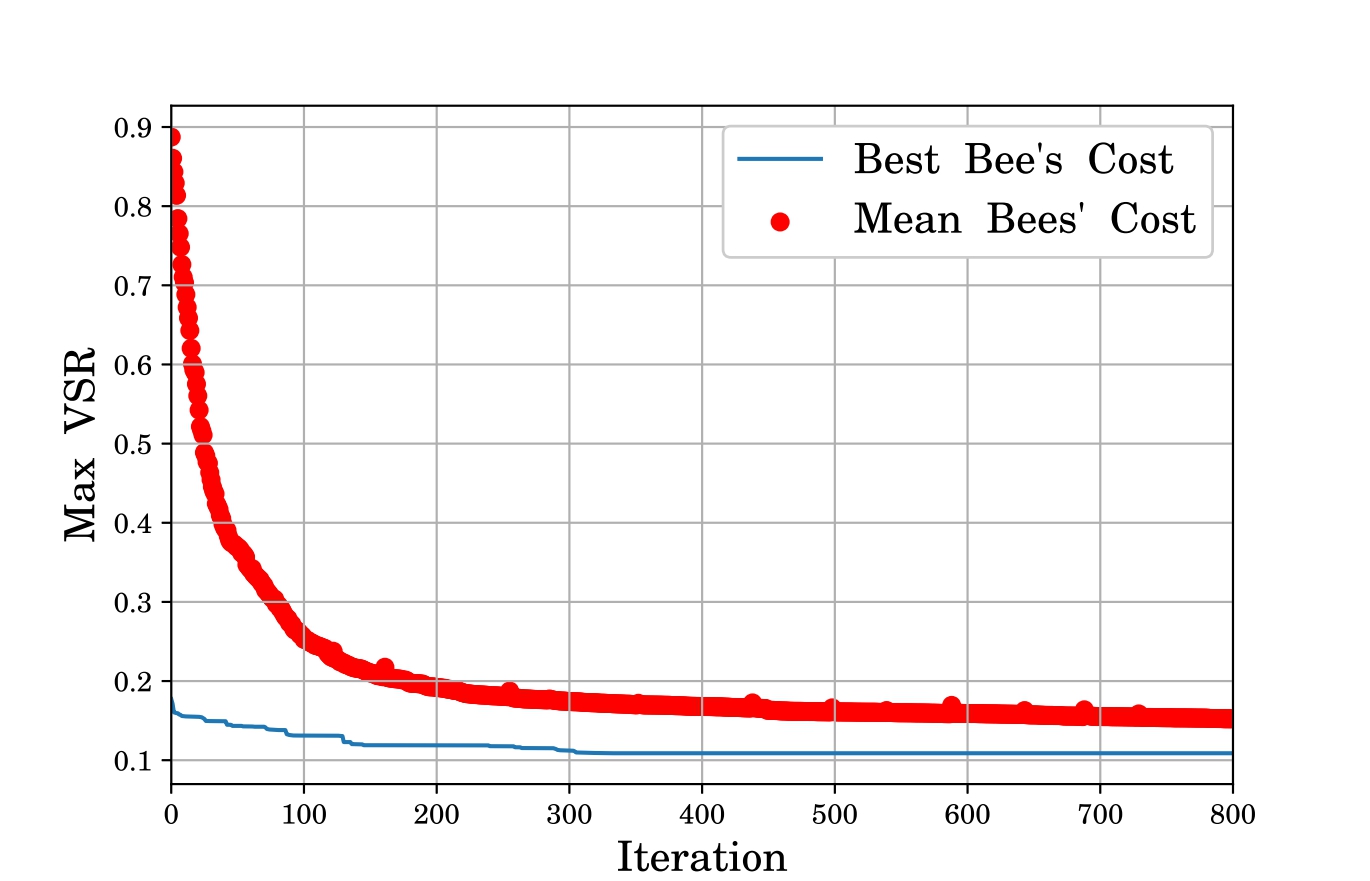} 
                \caption{The convergence trend of ABC solutions for minimizing the cost function - the max VSR}
                \label{fig:cost_trend}
        \end{figure}

\begin{figure} 
                \centering
                \subfigure[1 16-beam LiDAR]
                {
                        \label{fig:vis1} 
                        \includegraphics[width=1.55in]{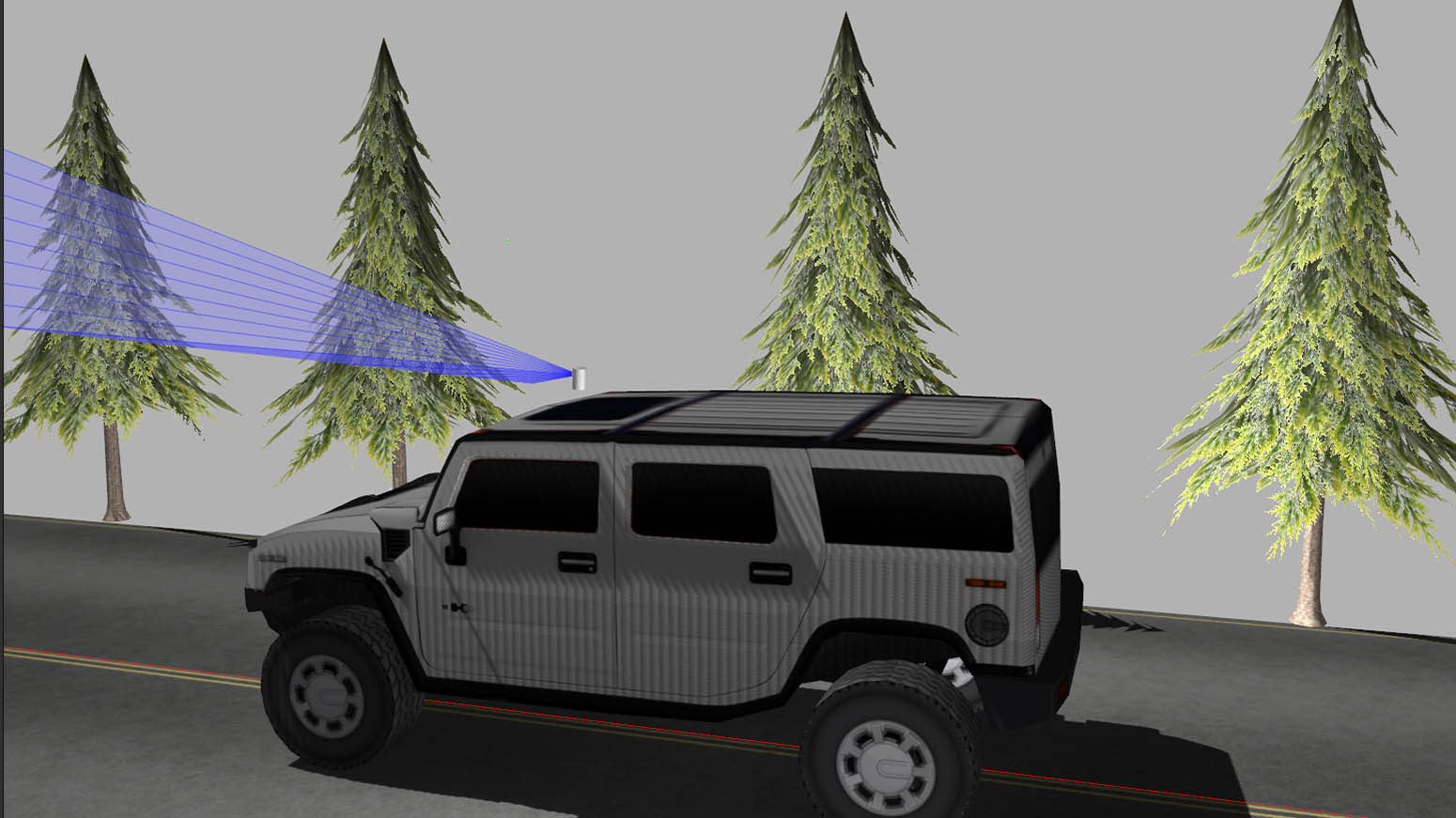}
                }
                \subfigure[2 16-beam LiDARs]
                { 
                        \label{fig:vis2}
                        \includegraphics[width=1.55in]{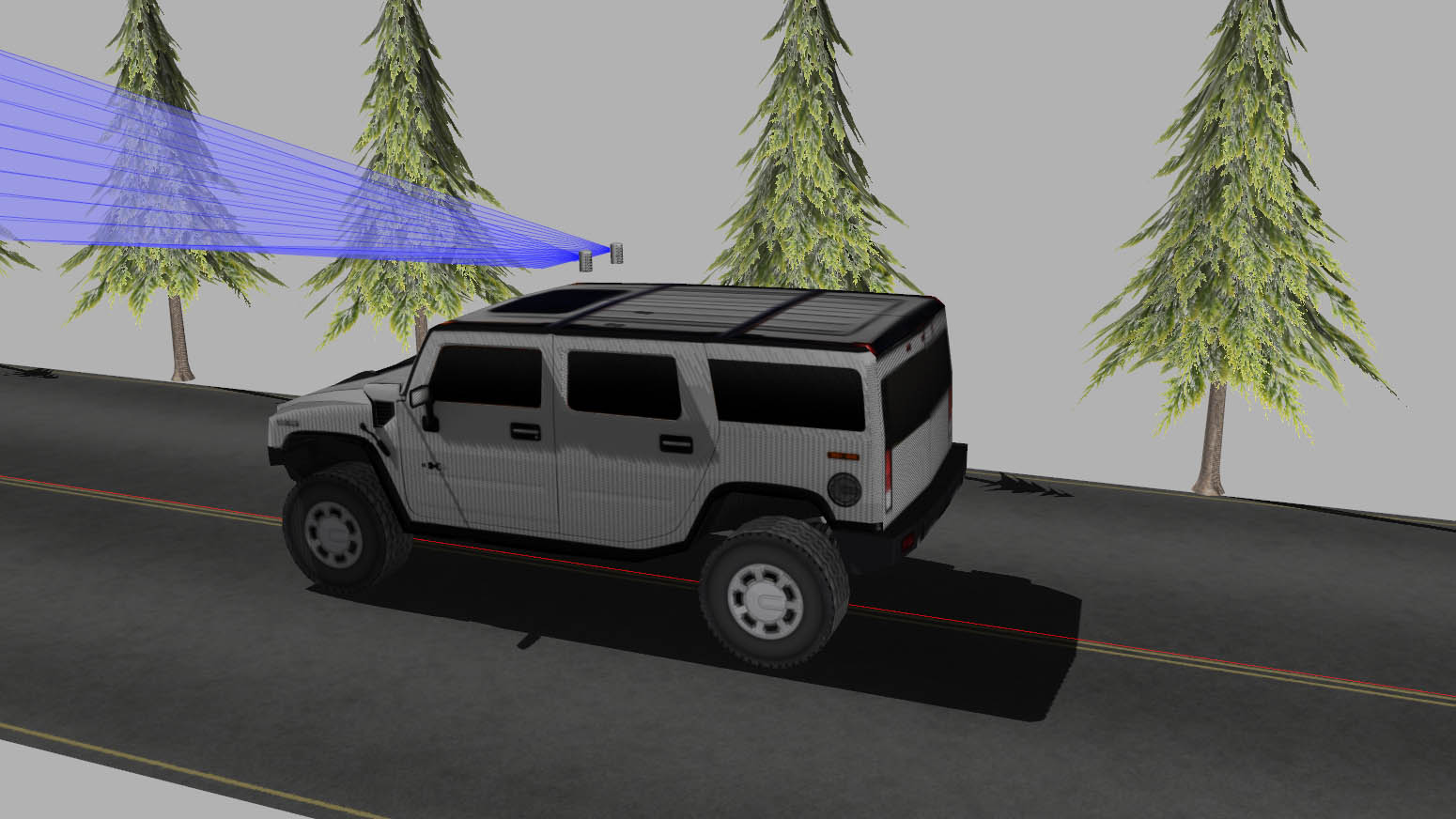}
                } 
                \subfigure[3 16-beam LiDARs]
                {
                        \label{fig:vis3} 
                        \includegraphics[width=1.55in]{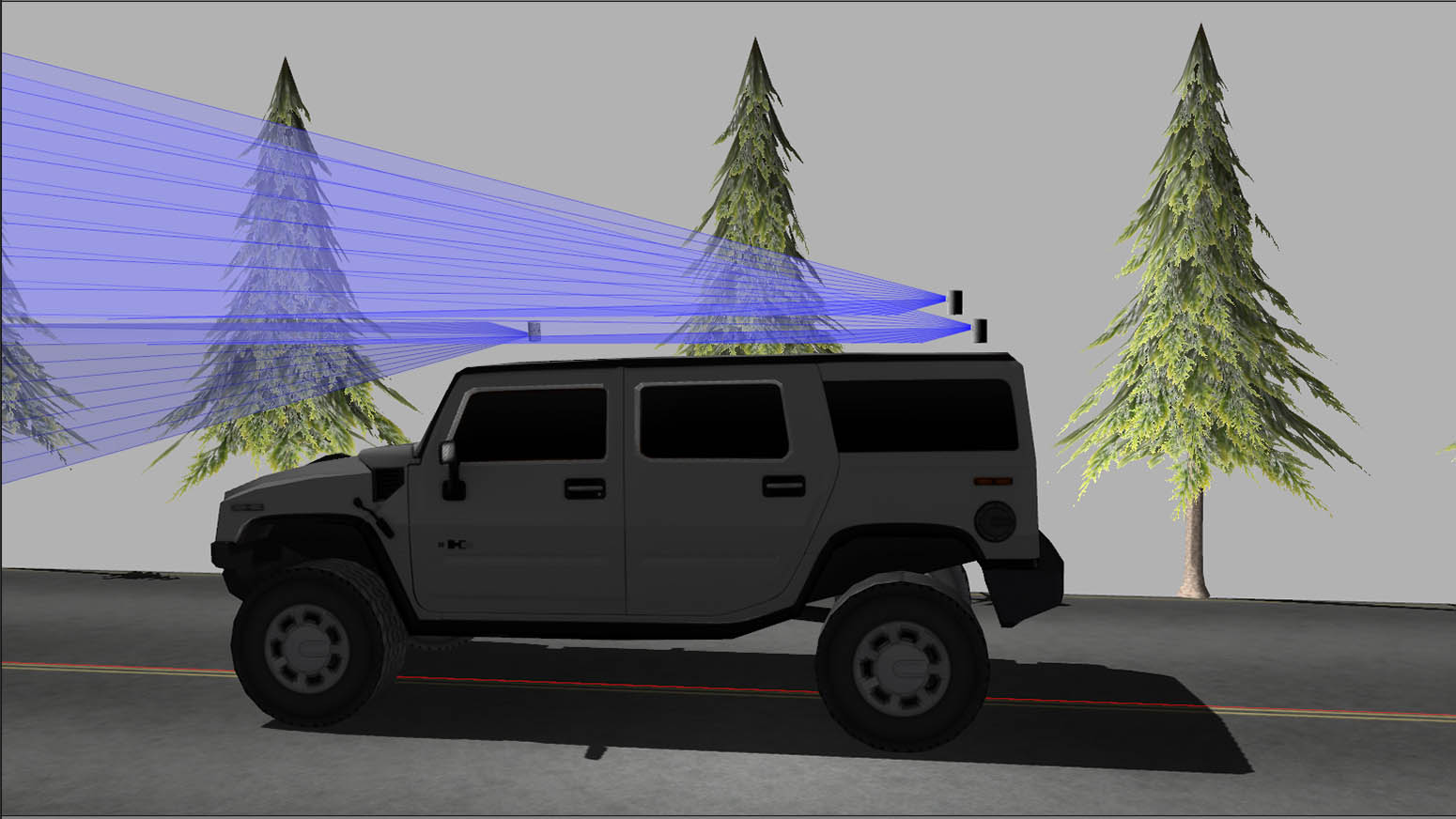}
                }
                \subfigure[4 16-beam LiDARs]
                { 
                        \label{fig:vis4}
                        \includegraphics[width=1.55in]{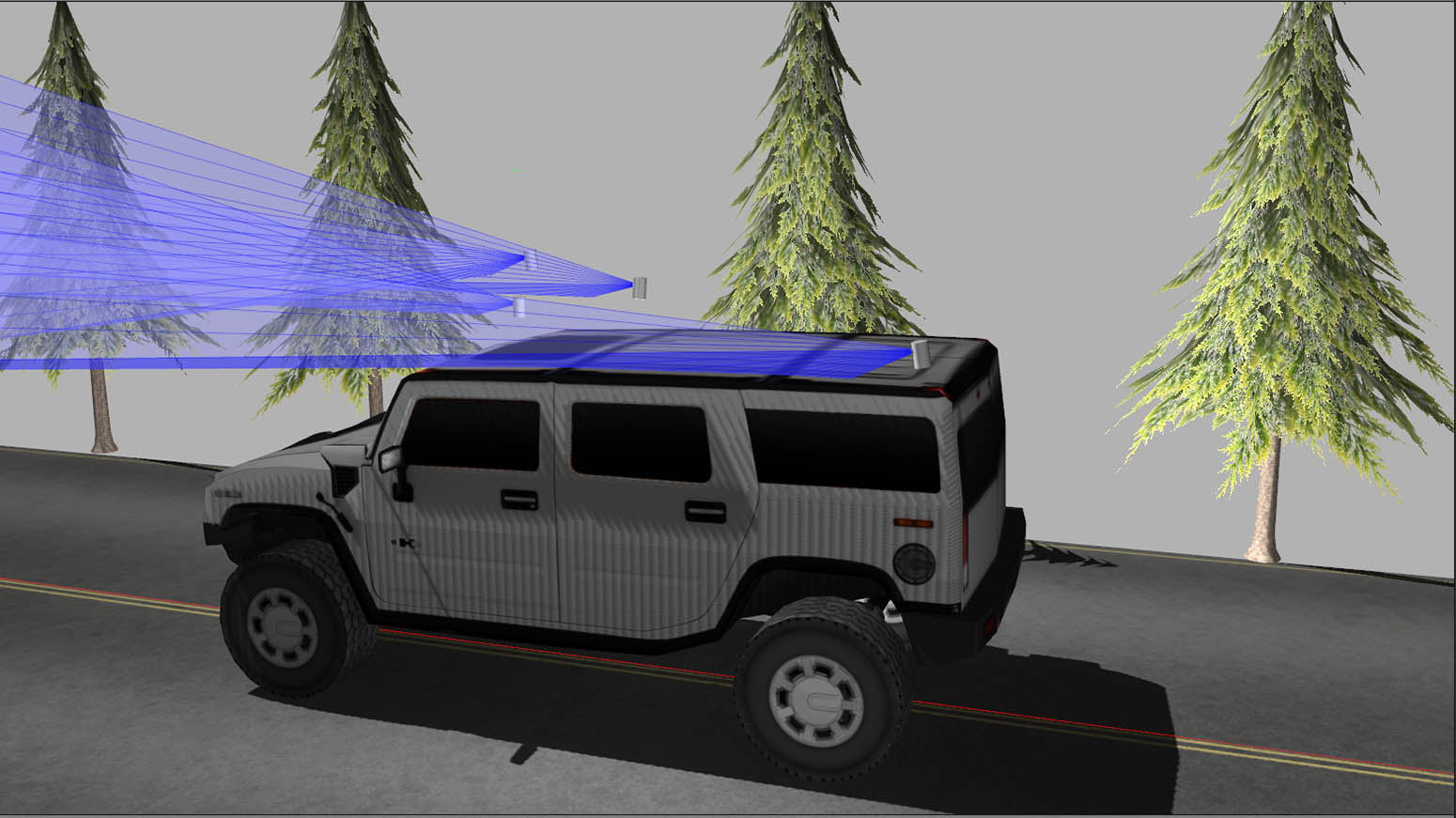}
                } 
                \caption{LiDAR configuration visulization}
                \label{vis}
        \end{figure}

        
        We use Velodyne VLP-16 \cite{velod16} as our candidate LiDAR components. For tractability, we set the resolution of the cuboids as $[1, 0.5, 0.2]$. To solve the problem, we set 200 bees and use the ABC algorithm to iterate 800 times. Fig. \ref{fig:cost_trend} shows how the ABC algorithm converges in the 4 LiDARs' case. The red line represents the average value of all the bees' max VSR, and the blue line represents the best solution's max VSR in each iteration. 
        Fig. \ref{fig:cost_trend} shows that the search area of the bees converges to the regions associated with lower cost values. Fig. \ref{vis} shows the optimized LiDAR configuration solutions for 1, 2, 3, and 4 LiDARs.

        
        
        
        \subsection{Maximum VSR trend in the number of LiDARs}
        Given the type of LiDAR, AV designers may need to determine a proper number of LiDARs. Fewer LiDARs may influence the detection performance, but too many LiDARs causes information redundancy and increases the computation burden. We showcase an application for determining the number and type of LiDARs to place. We consider three types of LiDARs: 4-beam, 8-beam, 16-beam LiDAR, with evenly distributed angles of the pitch from $-15^{\circ}$ to $15^{\circ}$. Our goal is to yield the optimal configuration and compare the detection uncertainty performance for the different types of LiDARs. 
        
        Fig. \ref{fig:best_cost} shows the results after 800 iterations. The fact that the marginal improvement rate decays as the number of LiDARs increase implies that the LiDAR beams’ space configuration, which can also be interpreted as the ability to detect small objects, may not increase when the number of LiDARs reaches a certain value. Optimal configuration plays a significant role when an AV design employs a few sensors, such as the artificial compound eye \cite{floreano2015science}. The fact that the maximum VSR of the AV design is either larger, or the same at best, when the configuration is not optimized, implies a dominated performance compared to the solution obtained by the proposed method.
        
        \subsection{Comparison between different LiDAR types}
\begin{figure}
                \centering
                \includegraphics[width=2.7in]{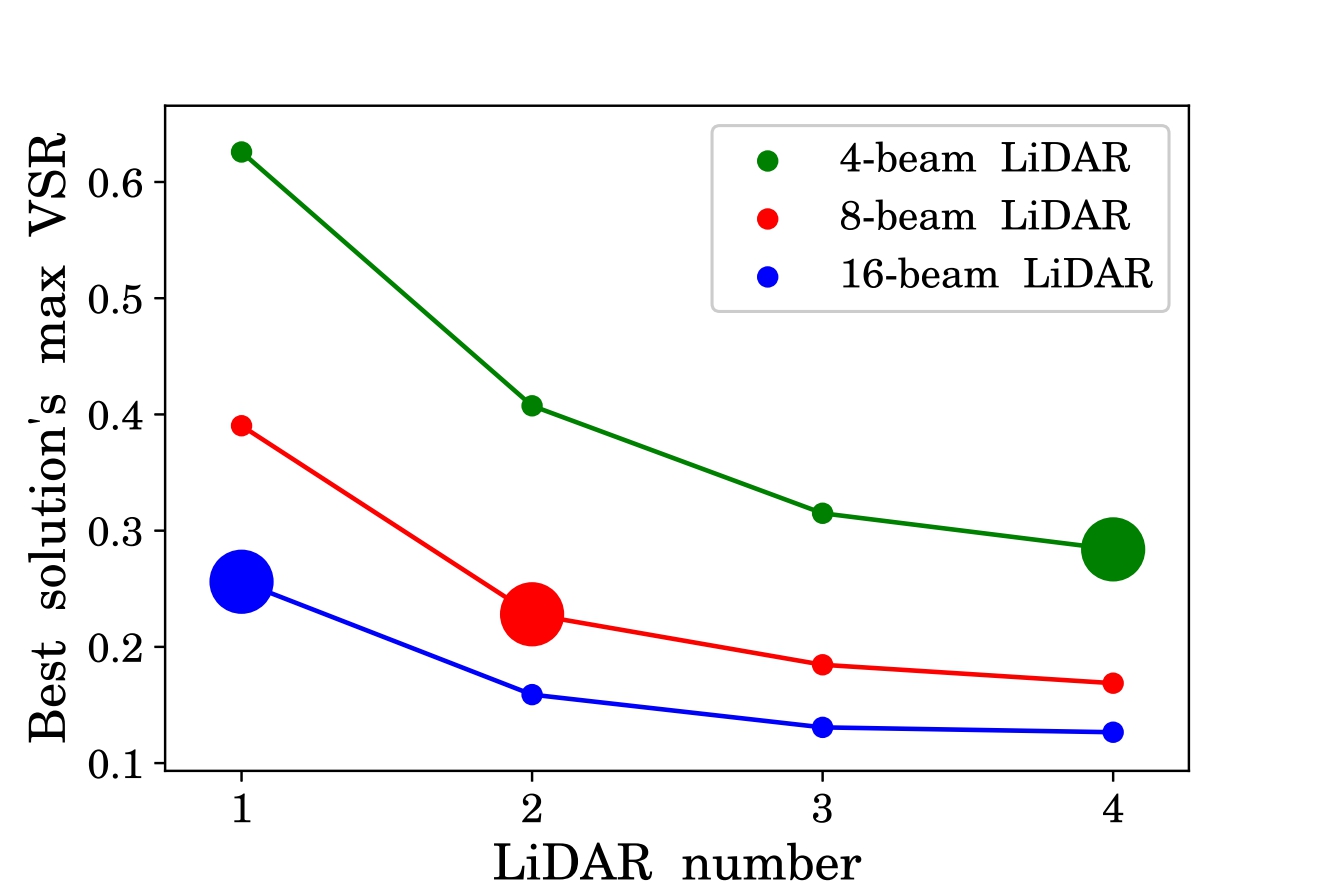}
                \caption{Best solution's maximum VSR of different LiDAR configuration after 800 iterations}
                \label{fig:best_cost}
        \end{figure}

        The proposed method can help AV engineers choose the optimal LiDAR types, or improve LiDAR design. For example, the marked points in Fig. \ref{fig:best_cost} show that two 8-beam LiDARs outperform both the one 16-beam and the four 4-beam LiDARs \textit{when configured optimally using our proposed method}. Intuitively, the four 4-beam LiDARs should be chosen, because its configuration is more flexible. However, its physical constraints, i.e., the angles of the pitch of its beams, can limit performance. This insight analysis will be highly advantageous for AV designer to better achieve affordable and safe AV design, and for LiDAR manufacturer better design the LiDAR type.


        \section{CONCLUSIONS}
        \label{sec:conclusion}
 
        This paper proposed a solution to the optimal LiDAR configuration problem which will help designers and manufacturers determine the proper number of sensors and their placement on autonomous vehicles. The overall relationship between the size of non-detectable subspaces, such as the blindspot of a given LiDAR configuration, the volume to surface area ratio (SVR), and the object detection rate (ODR) was investigated by using a new VSR-based measure. The Artificial Bee Colony (ABC) algorithm was adopted to segment the subspaces of the region of interest (ROI) into a cuboid-based representation. That yielded efficient computational performance. The experiment results indicated the effectiveness of the solution in prescribing configurations that obtained maximum detection performance without becoming computationally burdensome. 

        The research in this paper will be extended to investigate occlusion problems, and the solutions will be validated with real-world testing. The results are expected to bring AV designers, engineers, and manufacturers together to deploying autonomous vehicles and fleets with a reliable and affordable perception system.
        
        \addtolength{\textheight}{-12cm}        
        
        
        
        
        
        
        
        \bibliographystyle{IEEEtran}
        \bibliography{root}
        
        \end{document}